\pdfoutput=1

\documentclass[letterpaper,10pt,conference,final]{ieeeconf}
\IEEEoverridecommandlockouts
\usepackage{amsmath,amssymb,amsfonts}
\usepackage{algorithmic}
\usepackage{graphicx}
\usepackage{textcomp}
\usepackage{xcolor}
\usepackage{graphicx}
\usepackage{makecell}

\let\labelindent\relax
\usepackage{enumitem}

\usepackage[small]{caption}
\usepackage{booktabs}
\usepackage{hyperref}
\usepackage{comment}
\newtheorem{remark}{Remark}
\usepackage{float}
\usepackage{balance}

\usepackage{cleveref}

\usepackage[backend=bibtex,style=ieee,mincitenames=1,maxcitenames=2,maxbibnames=20]{biblatex}
\let\citet\textcite
\usepackage{hyperref}
\hypersetup{
    colorlinks=true,
    linkcolor=orange,
    filecolor=magenta,
    urlcolor=orange,
    citecolor=orange,
}
\usepackage{pdfx}

\title{\LARGE \bf Active Reward Learning from Online Preferences
\thanks{The authors would like to acknowledge the funding provided by the DARPA YFA award; NSF Awards \#2218760, \#1953032, and \#1941722; and the Office of Naval Research.}
}

\newenvironment{compact}[1][itemize]{
\itemsep=0pt
\partopsep=0pt
\topsep=0pt
\parskip=0pt
\parsep=0pt
\leftmargin=0pt
\csname#1\endcsname[leftmargin=*]
\def\done{end#1}
}{\expandafter\csname\done\endcsname}

\newenvironment{revision}{}{}

\author{Vivek Myers$^{1}$, Erdem B{\i}y{\i}k$^{2}$, Dorsa Sadigh$^{1,2}$ %
\thanks{$^1$Computer Science, Stanford University, $^2$Electrical Engineering, Stanford University. Emails: \texttt{vmyers}, \texttt{ebiyik}, \texttt{dorsa@cs.stanford.edu}}
}

\begin{document}

\maketitle

\begin{abstract}
Robot policies need to adapt to human preferences and/or new environments. Human experts may have the domain knowledge required to help robots achieve this adaptation. However, existing works often require costly offline re-training on human feedback, and those feedback usually need to be frequent and too complex for the humans to reliably provide. To avoid placing undue burden on human experts and allow quick adaptation in critical real-world situations, we propose designing and sparingly presenting easy-to-answer pairwise action preference queries in an online fashion. Our approach designs queries and determines when to present them to maximize the expected value derived from the queries' information. We demonstrate our approach with experiments in simulation, human user studies, and real robot experiments. In these settings, our approach outperforms baseline techniques while presenting fewer queries to human experts. Experiment videos, code and appendices are found on our website:\\ \url{http://tinyurl.com/online-active}
\end{abstract}

\edef\captionslist{ %
  	(i) Keep a posterior $\omega$ over the task (location of the dark blue tape square) conditioned on all past queries made to the human during the trajectory.
  	(ii) Consider potential preference queries $(a_1, a_2)$ between two actions.
  	(iii) Condition the posterior on each possible response to the query.
  	(iv) Use human model along with expected return of trajectories $\xi_1$ and $\xi_2$ under conditioned $\omega$ posteriors to compute value of query. Repeat for several query pairs of actions. 
  	(v) If the computed value of any potential query exceeds threshold, send the highest-value query considered to the human.
}

\section{Introduction}
Using trained robot policies in a zero-shot manner in real-world scenarios is challenging for many reasons, including insufficient training data, changing preferences of human users, or uncertainties present in dynamic environments. Today's robot algorithms need to fine-tune and adapt to the specifics of a given environment or user preferences in an online fashion. For example, the robot may need to safely explore the new environment and update its policy, but this is often too costly. An alternative is to require the human intervention by providing more data (such as demonstrations, physical feedback, language corrections)~\cite{li2021learning}, or by formally specifying new reward functions~\cite{hadfield2017inverse,ratner2018simplifying}, which might be too difficult and expensive. In addition, the common paradigm of learning from such feedback requires the robot to run its old policy, gather information during this run, pause to train on this data, update the policy based on the newly provided data, and to only then initiate a new run. This is highly impractical in online settings, as we already observe in interactive imitation learning~\cite{dagger,kelly2019hg,menda2019ensembledagger,hoque2021lazydagger,hoque2021thriftydagger}.

Instead, we would like the robot to reduce its uncertainty and update its policy by effectively asking humans questions and learning from their responses in an \emph{online fashion}.
This requires the robot to generate informative online questions that i) reduce the robot's uncertainty, ii) can easily be answered by humans, and iii) are critically timed, i.e., the robot should query humans infrequently and at the right time. 
Prior work has discovered pairwise comparisons as an effective form of feedback that can easily be answered by humans to update the robot's policy~\cite{wirth2017survey,christiano2017deep,cakmak2011human,li2021roial,biyik2020active,biyik2021learning}.
For instance, active preference-based learning techniques are data-efficient approaches that query humans with the most informative pairwise comparison---asking the human to compare two different robot trajectories---based on optimizing an information theoretic objective such as mutual information \cite{biyik2019asking}, volume removal \cite{sadigh2017active,biyik2018batch}, maximum regret \cite{wilde2020active,wilde2021learning}, dissimilarity \cite{katz2019learning,katz2021preference}, or determinantal point processes \cite{biyik2019batch}.
However, these works still require the policy to be updated during an offline training phase making them impractical for online policy updates. \begin{revision}The online setting we are interested in best captures many real-world scenarios in which a robot is deployed in a high-value situation where it is worth querying humans during the run to avoid poor performance. These scenarios include self-driving settings, where bothering a human is worth it to avoid an accident; high-level task planning settings where robots need efficient clarification about which task to perform next; and manipulation settings such as pushing, grasping, door opening, etc. where the human may have strong preferences about the precise task to perform\end{revision}. In such real-world settings, we additionally wish to avoid asking too many questions so that humans can quickly and reliably respond to questions when they are most needed.

\begin{figure*}[hbt]
\centering
    \vspace{10px}
  \includegraphics[width=.8\linewidth]{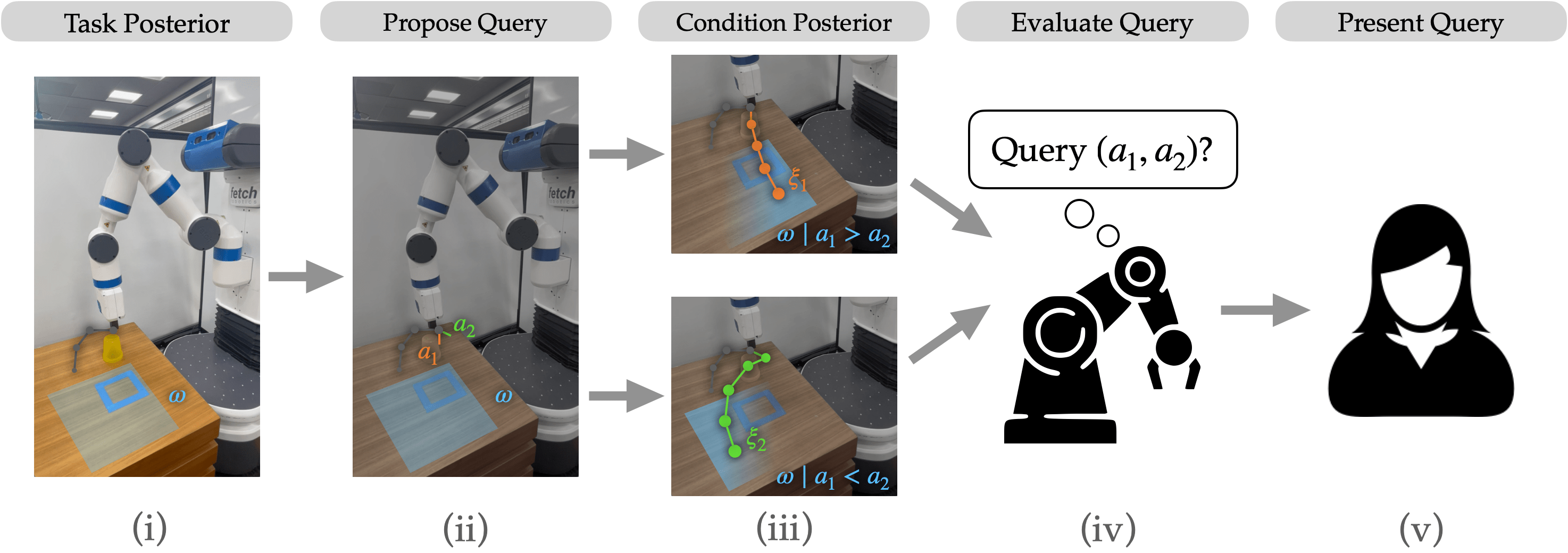}
  \captionsetup{singlelinecheck=off}
  \caption{
	  Overview of our approach to decide if and what to query each time step.
	  The robot must move the yellow cup to the goal location desired by the human, denoted by the blue tape square. 
	  The robot does not observe this goal location, and must query the human to determine where to move the cup.\captionslist
  }
  \label{fig:front}
\end{figure*}

This brings us to our key insight: we would like to model the robot's epistemic uncertainty -- its uncertainty about the environment -- captured by the robot's reward function. This uncertainty allows the robot to evaluate \emph{when} to ask \emph{the most informative} pairwise comparisons to the human.
Specifically, we take a multitask learning perspective to this problem and pretrain a library of robot reward functions on a number of tasks. In test time, we maintain a posterior over different tasks by directly modeling the effect of presenting queries to a human on the robot's belief state. We can thus compute the expected value of information (EVOI) \cite{viappiani2010optimal,cohn2011comparing} of any potential pairwise query corresponding to the expected value the robot gains by asking that question.
Using the EVOI metric, we propose an approach for selecting when to query a human expert and what queries to make. This is demonstrated through an example in \Cref{fig:front}: here, the robot uses our approach to query a human expert and push an object to a preferred goal location. When the expected value of a question is high enough, we query the pair of actions with the highest expected value of information.
Our approach asks pairwise comparison queries over the robot's actions (or the next short sequence of actions) that can be conveniently answered by humans. In addition, leveraging the EVOI metric allows us to ask the most informative questions at the critical time steps and thus update our policy in an online fashion.


\begin{revision}I\end{revision}n sum, we make the following contributions:

\vskip 1ex
\begin{compact}[enumerate]
	\item We propose a formulation of the online pairwise action preference querying problem using a multitask setting to model reward uncertainty, and show how to use this formulation to compute the value of asking questions.
	\item We show that our approach outperforms baselines by attaining a greater score while asking fewer questions in a tabular GridWorld game, a driving simulation, and continuous-action-space object-pushing robot environment.
	\item We conduct a user study with the driving simulation, showing users prefer the questions and the performance of our approach compared to baselines. We further demonstrate that our approach transfers to a real robot task allowing it to move an object to a preferred location.
\end{compact}

\section{Related Work}

Existing work on incorporating human feedback in robot learning has focused on interactive imitation learning as well as work on learning from other sources of data such as physical feedback, language instructions and corrections, rankings or pairwise comparisons. These works generally require extensive human feedback and/or learn from experts offline, whereas our approach is the only, to our knowledge, to both learn from humans using simple pairwise queries and perform that querying in an online fashion.

\textbf{Utilizing Human Experts Online.}
Most prior work on adaptive online querying of human experts has focused on requesting full demonstrations. DAgger requires rollouts from a policy that uses expert predictions \cite{dagger}, while ThriftyDAgger \cite{hoque2021thriftydagger} temporarily shifts control to human experts for partial demonstrations in risky or uncertain situations. Similarly, approaches such as uncertainty-aware action advising request demonstrations in high-uncertainty situations \cite{da2020uncertainty}. 
Building upon these interactive imitation learning work, HG-DAgger \cite{kelly2019hg} and EIL \cite{spencer2022expert} require human experts to decide when to stage interventions. These works require access to expert demonstrations which can be costly for robots with high degrees of freedom~\cite{argall2009survey,akgun2012keyframe,dragan2012formalizing,hindsight,khurshid2015data}, and further require a training phase based on the collected data.

To avoid excessively burdening humans experts by requiring full demonstrations, \citet{cohn2011comparing} leveraged EVOI and assumed the robot can move to any state to ask for the optimal action in that state. Approaches such as TAMER, COACH, and variants request continuous scalar feedback from human experts \cite{tamer,coach,dqntamer,deeptamer}.
Other works tap into other sources of data such as physical corrections, language instructions or corrections, images, and rankings~\cite{biyik2021learning,li2021learning,bajcsy2018learning,jamieson2020active}. While these works provide effective interfaces for collecting human data, they all require an offline training phase which does not allow for seamless online integration of new user inputs. 



\textbf{Preference-based Learning.}
Many works have examined reinforcement learning in the context of human preferences \cite{wirth2017survey}.
In particular, past work tackled reward learning in the context of preference-based queries \cite{biyik2020active,sadigh2017active,myers2022learning,biyik2018batch,biyik2019asking,tucker2020preference,christiano2017deep,basu2019active,hejna2022fewshot}. These works adaptively request human rankings or comparisons over trajectories to infer reward functions. 
\citet{habibian2022here} expand these techniques to ask questions that demonstrate to human experts that the robot is learning from their instruction, while \citet{biyik2019asking} and \citet{jeon2020reward} propose supplementing preferences with diverse forms of human feedback. \begin{revision}Meanwhile, within the student-teacher reinforcement learning framework, \citet{zimmer2014teacher} propose an approach where a teacher agent gives action preferences to a student agent to help it learn quickly.\end{revision}

Other approaches such as T-REX attempt to learn reward functions that allow effective policies to be trained from rankings of suboptimal trajectories \cite{trex}, while D-REX builds on this approach to learn from suboptimal trajectories without requiring expert rankings \cite{drex}.

Outside of reinforcement learning, preference-based active learning has been used for classification tasks \cite{chen2013pairwise,chen2017near} and ranking aggregation \cite{chen2013pairwise}. 
None of these approaches have tackled our problem of adaptively asking for pairwise comparisons between actions at deployment to conduct reward learning in an online setting.




\section{Learning When to Make Informative Queries}

\label{sec:approach}

We represent the robot's uncertainty as a distribution over possible reward functions. This distribution represents variability in human preferences about how a task should be performed. We assume our reward function in our environment is parameterized by a hidden latent task vector $\omega$. The robot does not have access to this hidden task vector during training or evaluation. However, it may during evaluation at any point ask the human for their preference between two actions. The human responds with their preference (up to some noise factor), assuming access to the true task vector. 

\textbf{Formalization.}
We consider a Markov Decision Process (MDP) of the form $(S,A,P,R^\omega)$ where $\omega$ is a task representation parameterizing the reward function $R^\omega$. We assume there is some distribution $\omega\sim\Omega$ which is known at train time and unknown at evaluation time. Having access to $\omega$ at train-time represents that we are able to learn different goal-conditioned policies during training, but must rely on human feedback to get information about the goal when deployed during evaluation.

\def\Q{Q^*}
Let us define $Q^*(s,a;\omega)$ to be the optimal $Q$-function for the MDP $(S,A,P,R^\omega)$ with optimal policy $\pi^*(s)=\arg\max_a Q^*(s, a)$. At evaluation-time, at a given state $s_t\in S$ at time $t$, we allow the robot to make an instantaneous action query of the form $(a_t^1, a_t^2) \in A \times A$, asking an expert to choose between the two possible actions.
Denote the state the agent is currently in as $s\in S$. \begin{revision}
We assume the human responses follow the Boltzmann-rational response model: \begin{equation}
   H^\pi(a_1,a_2,s,\omega)=\frac1{1+e^{\beta (Q^\pi(s, a_2;\omega) - Q^\pi(s, a_1;\omega))}}
     \label{eq:response}
\end{equation} 
with a fixed precision constant $\beta$ (higher values indicate a more accurate human response model). The human returns $a_1$ with probability $H^{\pi^*}(a_1,a_2,s,\omega)$ and returns $a_2$ otherwise with probability $H^{\pi^*}(a_2,a_1,s,\omega)$.\end{revision}
Assuming this response model for pairwise comparisons between action queries, our goal is to find out \emph{when} to ask \emph{what} questions to the user.
This translates to the objective of asking informative questions that \emph{maximize the reward $R^\omega$} at evaluation, while also \emph{minimizing the number of queries to the human expert}. We next formalize this joint optimization.

\subsection{Querying Action Preferences} 
\label{sec:querying}

\textbf{Training.} During training, we assume we have access to a distribution of training tasks $\Omega$, or samples from this distribution $\omega_1, \ldots, \omega_n \sim \Omega$. We train policies across the distribution $\Omega$ with a method that learns a $Q$-function, such as DQN \cite{mnih2013playing}. We thus obtain $Q$ functions that are conditioned on the hidden task vector as $Q^\pi(s, a; \omega)$.

\textbf{Querying.}
We propose an approach based on the expected value of information (EVOI) \cite{viappiani2010optimal} for determining when it is optimal to make queries. Our key insight is that using a family of task-parameterized $Q$-functions, we can approximate the expected value of asking a question to an expert. We then use a threshold $c$ over the computed EVOI to determine when to generate a query. Intuitively, the threshold $c$ determines the willingness of the robot asking a question.

\def\post{\Omega\mid \mathcal D}
\def\D{\mathcal D}
\def\t#1{^{(#1)}}
At evaluation, we maintain a posterior over the task vector $\omega$. 
Using this posterior, we can decide on which actions to take as well as compute the EVOI of asking a human expert for a preference between two actions of the form $(a_1,a_2)$. We initialize the posterior $\Omega$ using the prior from the environment if it is known, and otherwise model the prior as a uniform distribution over the sampled tasks we are given. We update this posterior in response to each successive query made to the human expert, using the human response model. Writing the past set of responses, query states, and queries as $\D=\Bigl(\bigl(a^{(1)}_{i_1}, s\t 1, (a^{(1)}_{1},  a^{(1)}_{2})\bigr),\bigl(a^{(2)}_{i_2}, s\t 2, (a^{(2)}_{1}, a^{(2)}_{2})\bigr),$
\,$\ldots, \bigl(a^{(n)}_{i_n},s\t n, (a^{(n)}_{1}, a^{(n)}_{2})\bigr)\Bigr)$ for each $i_\bullet \in \left\{1,2\right\}$, we obtain the following posterior update formula
 \begin{align}
    \Pr\left[\omega \mid \Omega,\D\right] \propto \Pr[\omega\mid\Omega]\prod_{j=1}^n H(a^{(j)}_{i_j},a^{(j)}_{3-i_j},s\t j,\omega).
  \label{eq:posterior}
\end{align}

We marginalize across this posterior at evaluation to obtain the following evaluation policy
    $\pi(s) \!=\! \arg\max_a \mathbb E_{\omega\mid\Omega,\mathcal D} Q^\pi(s, a; \omega)$.
To compute the EVOI, we calculate the expected improvement in the expected $Q$ value of the action taken by the robot, marginalizing across the $\omega$ posterior over the human goal belief conditioned on past query responses. The formulation selects queries based on how much they increase the value function evaluated on the current state, which we can model using our human response model and the effect of the queries on our posterior between reward functions.

One key assumption we make here in defining the EVOI is that the optimal policy given our belief over $\omega$ is well-represented by $Q$-values $\mathbb E_{\omega\mid\Omega,\mathcal D} Q^\pi(s, a; \omega)$. We make this assumption to avoid the computational burden of needing to refit policies for every possible distribution $\Omega\mid\D$ at each time step. This heuristic of setting the expected $Q$ value to the weighted $Q$ value of policies parameterized across tasks does result in optimistic myopic errors when optimal task policies will disagree substantially about which action to take in future.

A second assumption is made in the pessimistically myopic nature of this EVOI objective---the objective approximates the expected gain in value function at the current state when a query is asked assuming no further queries will be made during the current trajectory.

Intuitively, these two forms of myopic error may actually counteract each other. In cases where the $Q$ function approximation is overconfident due to the first assumption (the optimal task policies will disagree in the future), the EVOI for querying will likely be high anyways at those future points of disagreement. This results in the $Q$ value heuristic overestimate capturing the fact that the approach will do better than the myopic optimal policy that makes no more queries. In other words, the heuristic of using the expected $Q$ across tasks is sometimes overly optimistic, but in such cases the overconfidence stems from future states with high EVOI where uncertainty will be resolved anyways. In practice, we indeed find that this heuristic of using expected $Q$ performs well in our experimental settings.

\def\E{\mathop{\mathbb E}}
Denote $\D_1=\D\cup \left\{(a_1, s, (a_1, a_2))\right\}$ and $\D_2= \D\cup \left\{(a_2, s, (a_1, a_2))\right\}$.
We obtain the following formula for the EVOI of the query $(a_1,a_2)$. A complete derivation of this formula can be found in \Cref{app:derivation}.
\begin{align*}
    &\operatorname{EVOI}(a_1, a_2) = \\
    &\quad\E_{\omega\mid\Omega,\D} \Bigl[
    H^\pi(a_1, a_2, s, \omega) \mathop{\mathbb E}_{\omega'\mid\Omega,\D_1}\max_a\left[Q^\pi(s,a;\omega') \right] \\
    &\qquad + H^\pi(a_2, a_1, s, \omega) \!\mathop{\mathbb E}_{\omega'\mid\Omega,\D_2}\! \max_a\left[Q^\pi(s,a;\omega') \right] \\
    &\qquad - \max_a Q^\pi(s, a; \omega) \Bigr]
    \refstepcounter{equation}
    \tag\theequation
    \label{eq:evoi}
\end{align*}

When the EVOI of any query at the current state exceeds the query threshold $c$ we make the query with the highest EVOI and update the posterior appropriately.


\subsection{Continuous Action Spaces}
\label{sec:continuous}
In settings with a continuous action space, two issues arise: (1) computing \Cref{eq:evoi} for every pair of actions and (2) computing the $\max_a Q$ terms in \Cref{eq:evoi} for any action becomes intractable.

To solve (1), we randomly sample a fixed number of potential queries from the action space, and query the highest-EVOI of these queries if it exceeds $c$. To solve (2), we assume access to task-specific policies $\pi(s;\omega)\!=\!\arg\max_a Q^\pi(s, a; \omega)$, allowing us to compute $\max_a Q^\pi(s, a; \omega) \!=\! Q^\pi(s, \pi(s;\omega);\omega)$. Actor-critic approaches can be used to maintain an approximate $\pi(s;\omega)$ value \cite{konda1999actor,grondman2012survey}, allowing us to approximate \Cref{eq:evoi}.
In the continuous setting when not querying, we approximate the optimal action by using the policy $\pi(s) \!=\! \mathop{\mathbb E}_{\omega\mid\Omega,\D} \pi(s;\omega)$.

\textbf{Selecting Hyperparameters.}
To select the constant $\beta$, we pick a value consistent with the scale of the rewards in our environment that models our belief over the accuracy of the human experts in intuitively assessing the goal-conditioned optimal $Q$-functions.
To select $c$, we find a value that empirically yields the desired number of queries made by our agent either in simulation or by examining data from another source such as the replay buffer during training. 



\section{Experiments}
\label{sec:experiments}

We conduct a number of simulated experiments in a GridWorld\footnote{To focus more on the other experiments where we work with real human users or a real robot, we present the results of the GridWorld simulations in \Cref{app:grid}, which can be found on the \href{http://tinyurl.com/online-active}{website}.} and a driving environment, perform a user study where humans respond to action preference queries in the driving domain, and demonstrate our algorithm can be run on a real robot arm for a reaching task.

\textbf{Baselines.}
We compare our approach against 2 baselines.

\emph{Random.} We present a random baseline policy that queries the top two actions in terms of $Q$-value with a fixed probability. In continuous action-space settings, the second-best action is not well-defined, so we instead select random actions to query as well.

\emph{Uncertainty.} The uncertainty baseline uses a method similar to the novelty heuristic in ThriftyDAgger \cite{hoque2021thriftydagger} to decide when to query. In particular, we present a query when the novelty of the current state is high (the variance of the $Q$-value of the best action integrating across $\omega\mid\Omega, \mathcal D$ exceeds a threshold). Unlike ThriftyDAgger \cite{hoque2021thriftydagger}, we cannot request full demonstrations in our setting. We instead query the top two actions by expected $Q$-value when our approach decides to query. Again, in continuous action-space settings, the second-best action is not well-defined, so instead we select actions by the amount they reduce the uncertainty of the $Q$-value of the action being selected.


To compare approaches in simulation, we simulate an expert policy which assumes access to the task representation as a proxy for a human expert which can answer queries made by the robot. We hypothesize that our approach will outperform baseline approaches in terms of the tradeoff between the task performance and the number of queries.

\subsection{Driving Simulation Experiments}
\label{sec:drivexp}
We conduct experiments in a simulated driving environment \cite{highway-env} where the goal is to control a car driving on a crowded highway, and the action space consists of $5$ discrete actions: $A = \{\text{shift left}, \text{shift right}, \text{slower},\text{faster},\text{idle}\}$. The tasks (different $\omega$) in this setting correspond to different preferences over the desired lane and lane changes, speed range, acceleration, and following distance. We train our policies across these tasks using DQNs \cite{mnih2013playing,stable-baselines3}.

\begin{figure*}[htb]
    \centering
    \makebox[\linewidth][c]{
    \includegraphics[width=.4\linewidth]{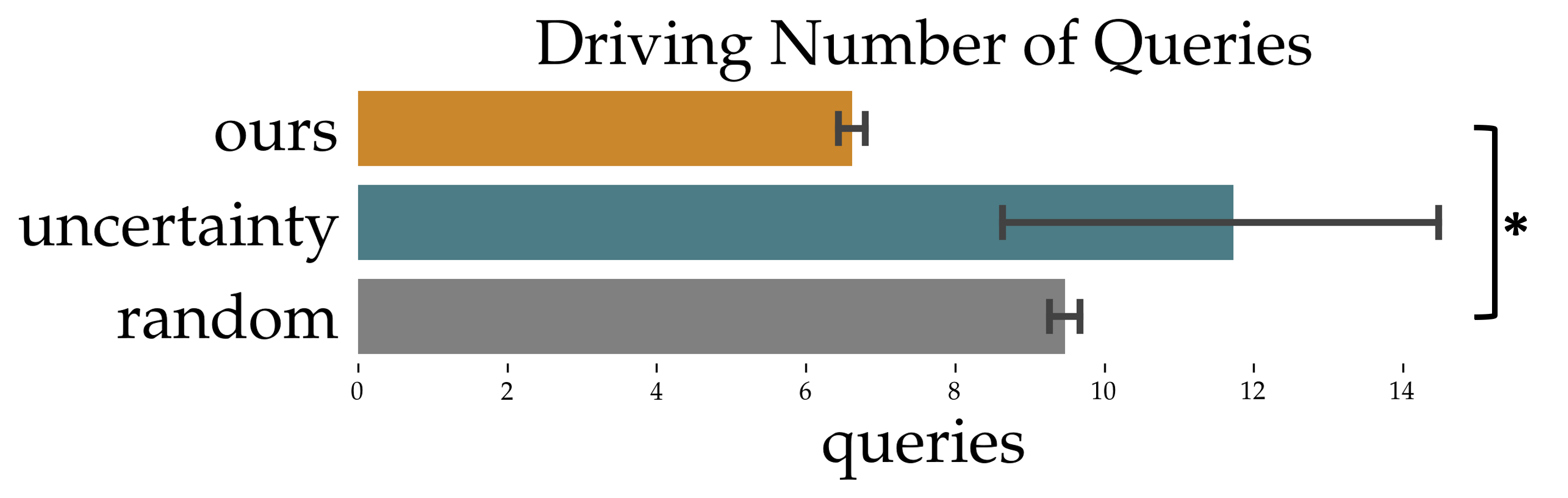}\hspace{10px}\includegraphics[width=.4\linewidth]{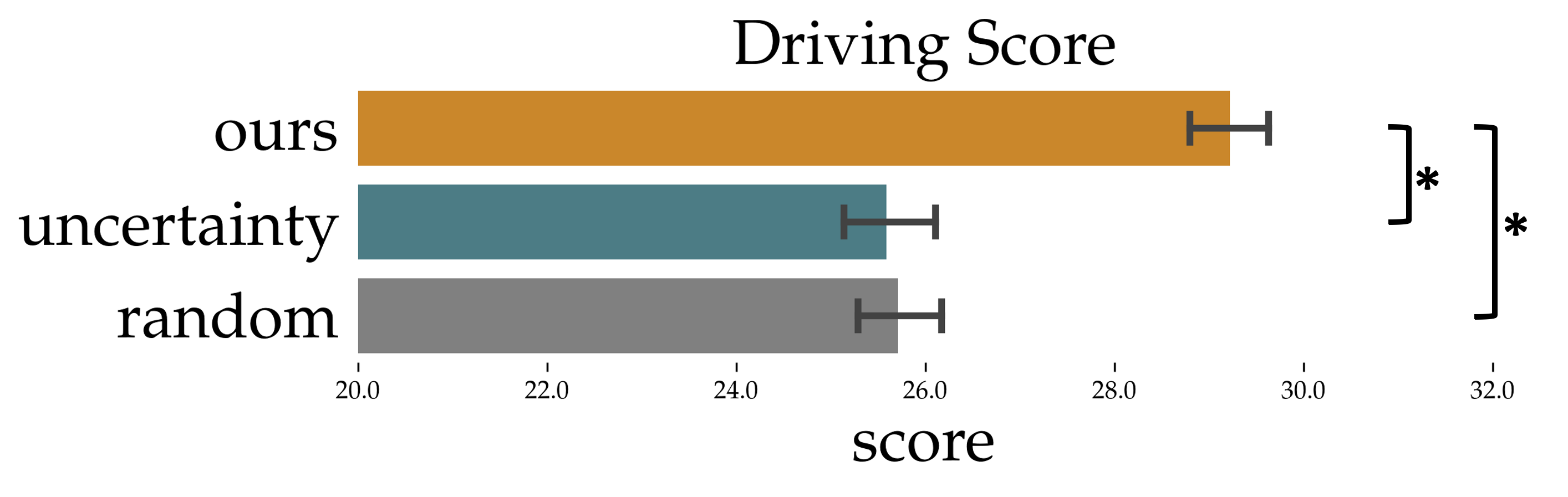}}
    \vspace{-15px}
    \caption{In the driving environment, our approach needs fewer queries on average than either baseline, while significantly outperforming both of them in average score across tasks.
    }
    \label{fig:driving}
\end{figure*}

\begin{figure*}[htb]
    \centering
    \includegraphics[width=0.8\linewidth]{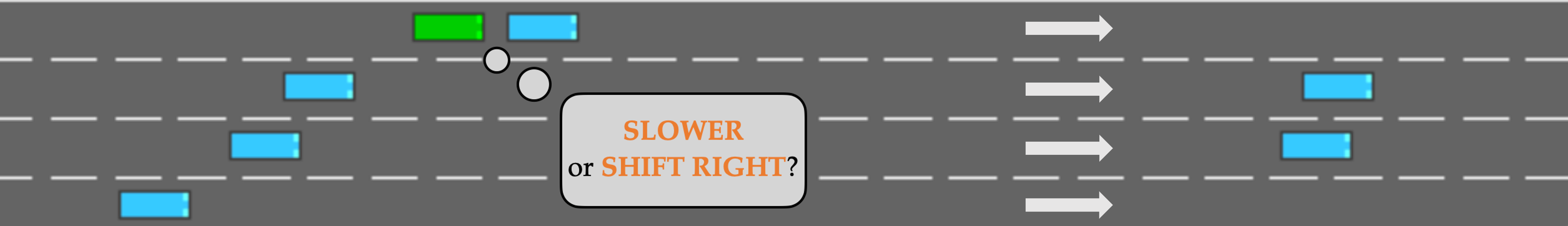}
    \caption{
    The ego car (green) is in a difficult situation where its desired speed is higher than the car in the front, but it cannot shift further left. A human with aggressive preferences may prefer to shift right one lane and try to overtake the car in front, while a less aggressive human would likely prefer slowing down. Since these different preferences dramatically affect the desirability of each approach, not knowing how aggressive the human driver is, our approach queries for their preference between slowing down and overtaking by turning right, updating its belief about the human based on their response.
    }
    \label{fig:thought}
\end{figure*}

We compared our approach against baselines in this environment across random different task initializations in \Cref{fig:driving}, using a simulated human expert modeled by a DQN with access to the task $\omega$. We tuned the $c$ parameter in simulation to ensure our approach and baselines made similar numbers of queries for fairness. An example of a decision made by our method is shown in \Cref{fig:thought}.

We then vary the querying threshold for each method (e.g., $c$ for ours) to analyze the tradeoff between number of queries and the performance (score) in the task. Pareto frontiers showing this tradeoff across different querying parameters are presented in \Cref{fig:pareto-combined}, demonstrating one can tune the $c$ constant of our method to set this tradeoff. These Pareto frontiers suggest our approach performs robustly and outperforms baselines.

\newcommand{\pmc}{\!\pm\!}
\setlength{\tabcolsep}{4pt}
\begin{table*}
    \centering
        \caption{User Study Results}
    \begin{tabular}{l|ccccc|cccc}
        \multicolumn{1}{c}{} & \multicolumn{5}{c}{Subjective Driving Metrics} & \multicolumn{4}{c}{Objective Driving Metrics}\\[1pt]
    \toprule
       Method & \thead{Important \\ Points} & \thead{Reasonable \\ Options}& \thead{Intelligent \\ Questions} & \thead{Adapted \\ Well} & \thead{Drove \\ Well}& \thead{Speed \\ Non-Adherence} & \thead{Lane \\ Adherence} & \thead{Number of \\ Queries} & \thead{Repetitive \\ Questions} \\
       \midrule
Ours  & $\mathbf{5.32 \pmc 0.26}$ & $\mathbf{4.95 \pmc 0.23}$ & $\mathbf{4.86 \pmc 0.18}$ & $\mathbf{5.32 \pmc 0.19}$ & $\mathbf{5.36 \pmc 0.20}$ & $\mathbf{7.94 \pmc 0.68}$     & $\mathbf{0.56 \pmc 0.10}$  & $\mathbf{8.24 \pmc 0.40}$  & $\mathbf{0.00 \pmc 0.00}$ \\
Uncertainty & $4.00\pmc0.42$ & $\mathbf{4.82\pmc0.36}$ & $\mathbf{4.27\pmc0.42}$ & $\mathbf{5.05\pmc0.39}$ & $\mathbf{4.73\pmc0.41}$ & $14.7\pmc2.2$ & $\mathbf{0.51\pmc0.08}$ & ${10.33\pmc0.31}$ & $8.51\pmc0.29$ \\
Random & $4.23\pmc0.30$ & $3.41\pmc0.31$ & $3.55\pmc0.33$ & $4.23\pmc0.32$ & $4.32\pmc0.36$ & $11.3\pmc1.6$ & $0.47\pmc0.05$ & $\mathbf{8.81\pmc0.33}$ & $1.03\pmc0.10$\\
\bottomrule
    \end{tabular}
    \label{tab:driving_results}
    \vspace{-10px}
\end{table*}

\subsection{Driving User Studies}
\label{sec:user_study}
With IRB approval from Stanford University Research Compliance Office, we repeated the driving experiments on real humans instead of simulated experts.

\textbf{Experimental Setup.}
We conducted our experiment using an online web interface. Subjects completed a pre-experiment survey in which they stated their gender, preferred lane, and preferred speed (between $20$ m/s and $30$ m/s). We additionally asked the subjects to state whether they preferred their car to adhere to their speed preferences or lane preferences more. Subjects were told to express consistent preferences, and, at the end, were asked to subjectively rate the effectiveness of each algorithm.

\textbf{Procedure.}
We gathered data from 22 subjects (11 female, 10 male, 1 preferred not to answer), and each subject responded to queries online over the course of 5 trajectories for each algorithm (our approach or one of the two baselines, unknown to the subjects) in randomized order.

\textbf{Independent Variables.} The querying method used, either our approach or one of the two baselines, is the independent variable.

\textbf{Dependent Measures.} Subjective measures are the evaluations performed by the subjects after each algorithm: a seven-point scale survey of how much they agree with the statements: ``The car asked questions at the important decision points," ``The questions included choices between reasonable options," ``Overall, the questions seemed intelligently timed and picked," ``Over time, the car adapted to my driving preferences," and ``Overall, the car drove in the way I wanted."

Objective measures include the speed non-adherence, the percentage relative difference between the average speed of the car and the subject's preferred speed; lane adherence, the proportion of the time the car stayed in the subject's preferred lane; as well as the number of queries needed by the algorithm and the number of repetitive questions asked, defined as consecutive steps where the algorithm asked the same question. Since adhering to speed preferences and lane preferences often conflict (when strongly adhering to a lane, speed is strongly determined by the other cars in that lane), we only included the subjects who preferred speed or lane adherence in the respective metric computations.


\textbf{Hypotheses.} (1) Our approach will more efficiently and better capture user preferences than baselines, indicated by obtaining higher scores on the subjective dependent measures. (2) Our approach will perform objectively better than baselines, indicated by greater speed adherence and lane adherence, and also quantitatively be more query efficient, indicated by using a lower number of queries and making fewer repetitive queries.


\textit{Results.}
\begin{revision}
\Cref{tab:driving_results} shows the different subjective assessments of the various approaches given by users with standard errors. \begin{revision}The best performing approaches for each metric (highest or lowest contextually) up to statistical significance are bolded (Wilcoxon signed-rank with a $p$-value of $0.05$).
\end{revision} Our approach achieves the best performance across all metrics, and is statistically significantly better for every metric in comparison to random, and statistically significantly better compared to uncertainty for the ``Important Points" metric, supporting \textbf{Hypothesis 1}.
\Cref{tab:driving_results} also shows the different objective assessments with standard errors. The best approaches are again bolded similarly. Our approach achieves the best performance across all metrics, supporting \textbf{Hypothesis 2}.
\end{revision}

\subsection{Real Robot Block Pushing Experiments}
\label{sec:pushing}
\begin{figure*}[t]
    \centering
    \makebox[.8\linewidth][c]{
    \includegraphics[width=.4\linewidth]{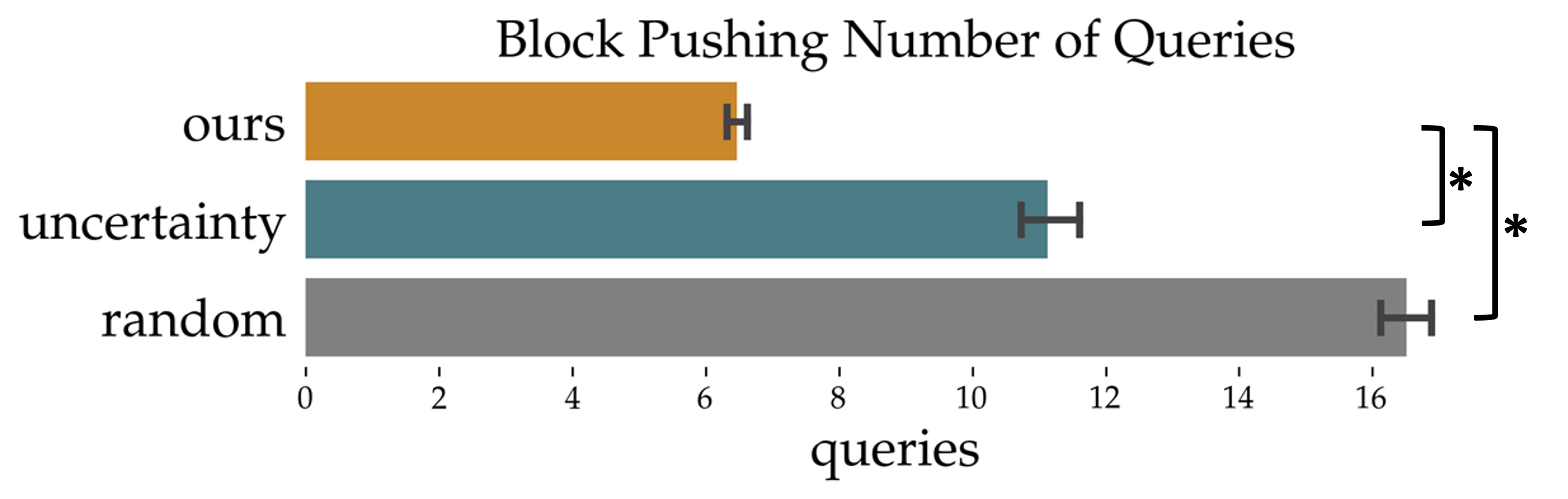}\hspace{10px}\includegraphics[width=.4\linewidth]{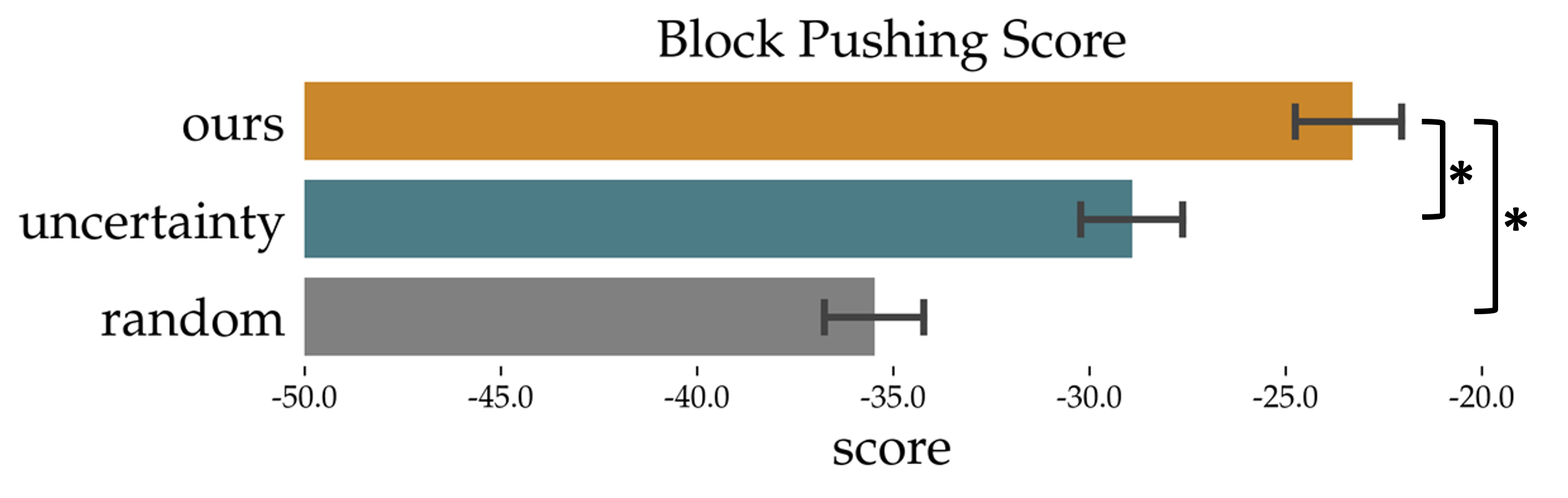}}
    \caption{Our approach makes statistically significantly fewer queries on average than all the baseline approaches in block-pushing simulation, while significantly outperforming them in average score across tasks. 
    }
    \label{fig:pushing}
\end{figure*}

\begin{figure*}[th]
  \centering
  \includegraphics[width=.8\linewidth]{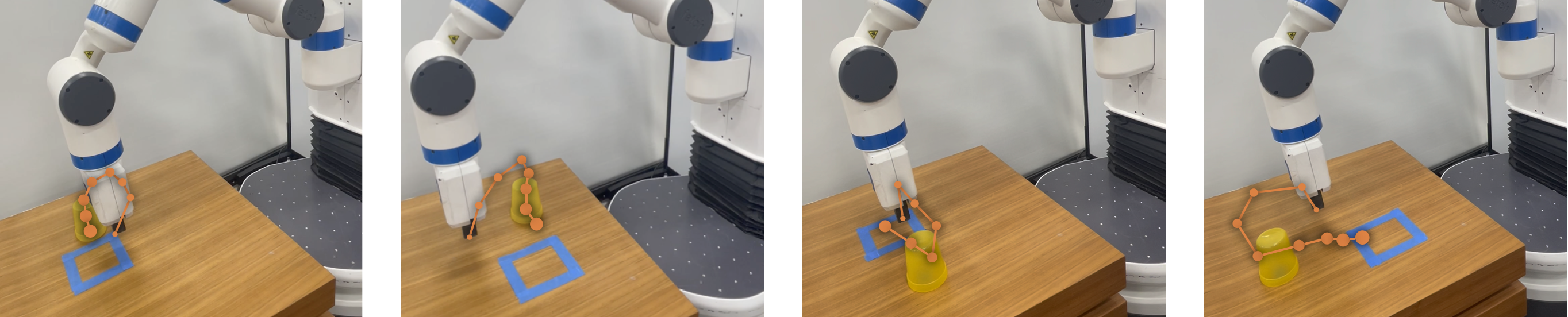}
  \caption{Example trajectories generated by our method on the Fetch robot with a simulated human expert. The goal location (blue tape square) is observed by the simulated expert but not our approach.}
  \label{fig:sim2real}
  \vspace{-10px}
\end{figure*}

\begin{figure}[H]
    \centering
    \includegraphics[width=\linewidth]{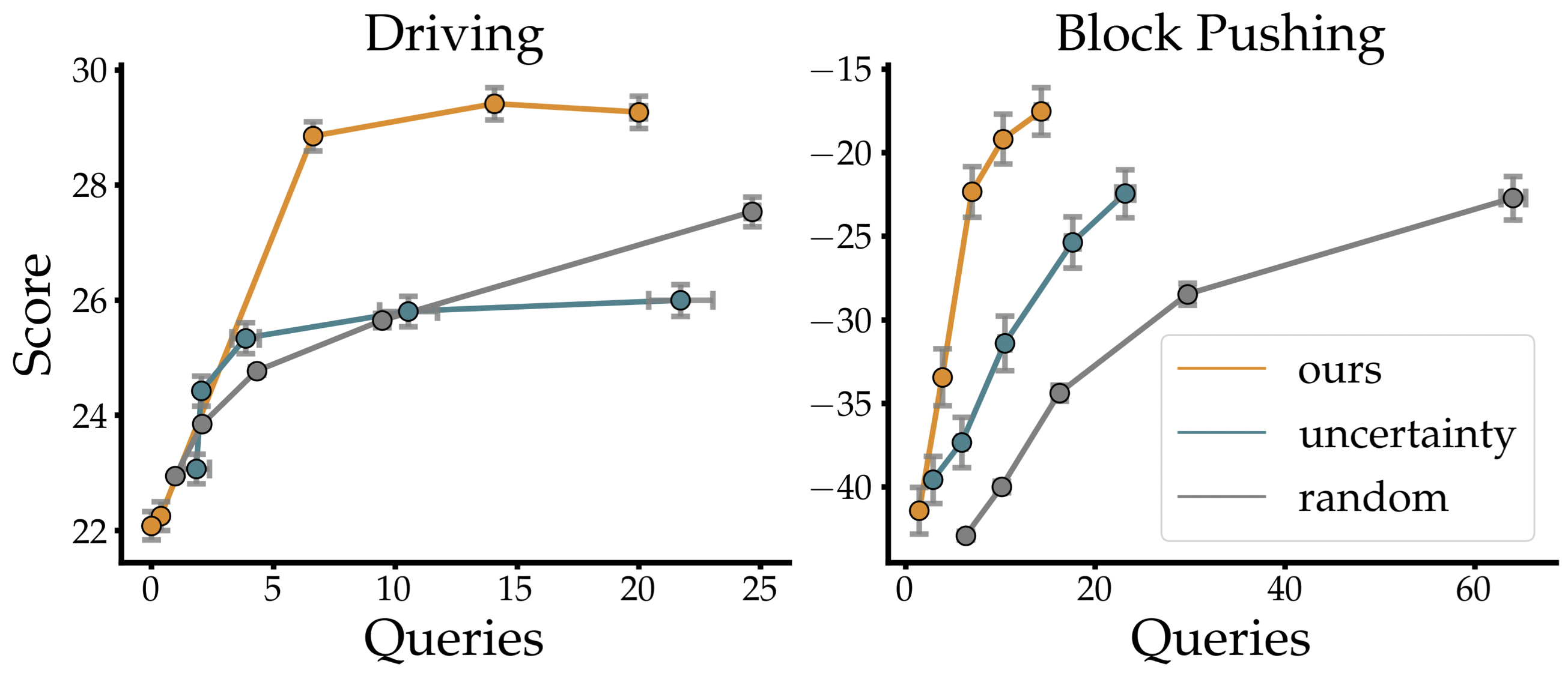}
    \caption{Pareto frontiers comparing approaches for the driving and block pushing environment. Error bars are one standard error.}
    \label{fig:pareto-combined}
\end{figure}

We first consider a simulated environment where a robot arm must be used to push a block to a goal location. We view the block location as the task $\omega$ in this setting, so to succeed in this setting, the robot must query the expert to gain information about the goal location. Since the action space is continuous, we use a trained soft actor--critic policy \cite{haarnoja2018soft,stable-baselines3} as described in \Cref{sec:querying}. We conducted experiments in simulation using simulated expert policies, and found that our approach outperformed baselines in score while needing fewer queries (see \Cref{fig:pushing}). We additionally constructed Pareto frontiers in \Cref{fig:pareto-combined} modeling the tradeoff between queries made and scores achieved for all methods, similar to driving simulations. These again show our method robustly outperforms baselines when asking similar numbers of questions. See \Cref{app:fetch} for the experiment details on the block pushing setting.

\textbf{Sim2Real Transfer.}
We then ran the approach on a real Fetch robot arm \cite{wise2016fetch}, using policies trained in the simulation. On initializations where our method succeeded in simulation, we were also able to transfer to the real robot and push the yellow cup to an arbitrary goal location, as shown in \Cref{fig:sim2real}. Further details about this experiment and the videos are on the \href{http://tinyurl.com/online-active}{website}.

\section{Discussion}
\begin{revision}

\textbf{Summary.}
In this paper, we proposed an approach that designs easy-to-answer pairwise action preference queries in an online fashion. Our work maximizes the expected value derived from the queries' information and asks questions from humans at the most informative points of time. We demonstrated our approach in simulated GridWorld, a driving simulation, and a pushing task with a real robot. Across all these settings our approach outperformed baselines while needing fewer questions. 




\textbf{Limitations and Future Work.}
One limitation of this work is the potential difficulty of answering preferences between actions. While the human response model in \Cref{eq:response} accounts for uncertainty in distinguishing similarly-valued actions, there is a challenge in effectively representing queries for humans in high DoF action spaces. In discrete settings where actions are easily interpretable, this is not an issue. In continuous settings such as most robot manipulation tasks, it can be harder for humans to interpret and compare actions without good representations---a potential solution is to use a simulator to present visualizations of both actions to human experts. In future work, we plan to filter queries to ensure they are easily answerable with a visualization.

An additional limitation is the assumption that queries can be immediately answered. In settings that require fast responses such as urban driving environments this assumption may be difficult. A future approach to tackle this problem would be to train a dynamics model and use this model to predict EVOI and design queries multiple steps in advance, giving experts more time to respond. 


In the future, we plan to apply this approach to more real-world robot tasks, such as Meta-World \cite{yu2020meta}. Additional exploration of complex simulation environments and baseline approaches would also help to validate our method.

Finally, theoretical analysis should be conducted on optimality, in particular with regard to (1) the effect of the approximations made to extend to continuous settings in \Cref{sec:continuous}, (2) the selection of $c$ and $\beta$ hyperparameters, and (3) the competing effects of the pessimistic and optimistic myopia of our approach discussed in \Cref{sec:querying}.

\end{revision}


\balance
\renewcommand*{\bibfont}{\small}
\printbibliography

@book{dragan2012formalizing,
  title={Formalizing assistive teleoperation},
  author={Dragan, Anca D and Srinivasa, Siddhartha S},
  year={2012},
  publisher={MIT Press, July}
}

@article{khurshid2015data,
  title={Data-driven motion mappings improve transparency in teleoperation},
  author={Khurshid, Rebecca P and Kuchenbecker, Katherine J},
  journal={Presence},
  volume={24},
  number={2},
  pages={132--154},
  year={2015},
  publisher={MIT Press}
}

@inproceedings{tamer,
  title={Tamer: Training an agent manually via evaluative reinforcement},
  author={Knox, W Bradley and Stone, Peter},
  booktitle={2008 7th IEEE international conference on development and learning},
  pages={292--297},
  year={2008},
  organization={IEEE}
}

@inproceedings{jamieson2020active,
  title={Active Reward Learning for Co-Robotic Vision Based Exploration in Bandwidth Limited Environments},
  author={Jamieson, Stewart and How, Jonathan P and Girdhar, Yogesh},
  booktitle={2020 IEEE International Conference on Robotics and Automation (ICRA)},
  pages={1806--1812},
  year={2020},
  organization={IEEE}
}

@inproceedings{coach,
  title={COACH: Learning continuous actions from corrective advice communicated by humans},
  author={Celemin, Carlos and Ruiz-del-Solar, Javier},
  booktitle={2015 International Conference on Advanced Robotics (ICAR)},
  pages={581--586},
  year={2015},
  organization={IEEE}
}

@article{plappert2018multi,
  title={Multi-goal reinforcement learning: Challenging robotics environments and request for research},
  author={Plappert, Matthias and Andrychowicz, Marcin and Ray, Alex and McGrew, Bob and Baker, Bowen and Powell, Glenn and Schneider, Jonas and Tobin, Josh and Chociej, Maciek and Welinder, Peter and others},
  journal={arXiv preprint arXiv:1802.09464},
  year={2018}
}

@inproceedings{deeptamer,
  title={Deep tamer: Interactive agent shaping in high-dimensional state spaces},
  author={Warnell, Garrett and Waytowich, Nicholas and Lawhern, Vernon and Stone, Peter},
  booktitle={Proceedings of the AAAI conference on artificial intelligence},
  volume={32},
  number={1},
  year={2018}
}

@article{dqntamer,
  title={Dqn-tamer: Human-in-the-loop reinforcement learning with intractable feedback},
  author={Arakawa, Riku and Kobayashi, Sosuke and Unno, Yuya and Tsuboi, Yuta and Maeda, Shin-ichi},
  journal={arXiv preprint arXiv:1810.11748},
  year={2018}
}

@inproceedings{zimmer2014teacher,
  title={Teacher-student framework: a reinforcement learning approach},
  author={Zimmer, Matthieu and Viappiani, Paolo and Weng, Paul},
  booktitle={AAMAS Workshop Autonomous Robots and Multirobot Systems},
  year={2014}
}

@inproceedings{basu2019active,
  title={Active learning of reward dynamics from hierarchical queries},
  author={Basu, Chandrayee and B{\i}y{\i}k, Erdem and He, Zhixun and Singhal, Mukesh and Sadigh, Dorsa},
  booktitle={2019 IEEE/RSJ International Conference on Intelligent Robots and Systems (IROS)},
  pages={120--127},
  year={2019},
  organization={IEEE}
}

@article{ratner2018simplifying,
  title={Simplifying reward design through divide-and-conquer},
  author={Ratner, Ellis and Hadfield-Menell, Dylan and Dragan, Anca D},
  journal={arXiv preprint arXiv:1806.02501},
  year={2018}
}

@article{argall2009survey,
  title={A survey of robot learning from demonstration},
  author={Argall, Brenna D and Chernova, Sonia and Veloso, Manuela and Browning, Brett},
  journal={Robotics and autonomous systems},
  volume={57},
  number={5},
  pages={469--483},
  year={2009},
  publisher={Elsevier}
}

@article{katz2021preference,
  title={Preference-based learning of reward function features},
  author={Katz, Sydney M and Maleki, Amir and B{\i}y{\i}k, Erdem and Kochenderfer, Mykel J},
  journal={arXiv preprint arXiv:2103.02727},
  year={2021}
}

@inproceedings{wilde2021learning,
 title={Learning Reward Functions from Scale Feedback},
 author={Wilde, Nils and Biyik, Erdem and Sadigh, Dorsa and Smith, Stephen L.},
 booktitle={Proceedings of the 5th Conference on Robot Learning (CoRL)},
 year={2021}
}

@inproceedings{cohn2011comparing,
  title={Comparing action-query strategies in semi-autonomous agents},
  author={Cohn, Robert and Durfee, Edmund and Singh, Satinder},
  booktitle={Proceedings of the AAAI Conference on Artificial Intelligence},
  volume={25},
  number={1},
  pages={1102--1107},
  year={2011}
}

@inproceedings{katz2019learning,
  title={Learning an urban air mobility encounter model from expert preferences},
  author={Katz, Sydney M and Le Bihan, Anne-Claire and Kochenderfer, Mykel J},
  booktitle={2019 IEEE/AIAA 38th Digital Avionics Systems Conference (DASC)},
  pages={1--8},
  year={2019},
  organization={IEEE}
}

@inproceedings{hoque2021lazydagger,
  title={Lazydagger: Reducing context switching in interactive imitation learning},
  author={Hoque, Ryan and Balakrishna, Ashwin and Putterman, Carl and Luo, Michael and Brown, Daniel S and Seita, Daniel and Thananjeyan, Brijen and Novoseller, Ellen and Goldberg, Ken},
  booktitle={2021 IEEE 17th International Conference on Automation Science and Engineering (CASE)},
  pages={502--509},
  year={2021},
  organization={IEEE}
}

@inproceedings{menda2019ensembledagger,
  title={Ensembledagger: A bayesian approach to safe imitation learning},
  author={Menda, Kunal and Driggs-Campbell, Katherine and Kochenderfer, Mykel J},
  booktitle={2019 IEEE/RSJ International Conference on Intelligent Robots and Systems (IROS)},
  pages={5041--5048},
  year={2019},
  organization={IEEE}
}

@inproceedings{li2021learning,
  title={Learning human objectives from sequences of physical corrections},
  author={Li, Mengxi and Canberk, Alper and Losey, Dylan P and Sadigh, Dorsa},
  booktitle={2021 IEEE International Conference on Robotics and Automation (ICRA)},
  pages={2877--2883},
  year={2021},
  organization={IEEE}
}

@article{hadfield2017inverse,
  title={Inverse reward design},
  author={Hadfield-Menell, Dylan and Milli, Smitha and Abbeel, Pieter and Russell, Stuart J and Dragan, Anca},
  journal={Advances in neural information processing systems},
  volume={30},
  year={2017}
}

@article{christiano2017deep,
  title={Deep reinforcement learning from human preferences},
  author={Christiano, Paul F and Leike, Jan and Brown, Tom and Martic, Miljan and Legg, Shane and Amodei, Dario},
  journal={Advances in neural information processing systems},
  volume={30},
  year={2017}
}

@inproceedings{tucker2020preference,
  title={Preference-based learning for exoskeleton gait optimization},
  author={Tucker, Maegan and Novoseller, Ellen and Kann, Claudia and Sui, Yanan and Yue, Yisong and Burdick, Joel W and Ames, Aaron D},
  booktitle={2020 IEEE international conference on robotics and automation (ICRA)},
  pages={2351--2357},
  year={2020},
  organization={IEEE}
}

@article{habibian2022here,
  title={Here’s what I’ve learned: Asking questions that reveal reward learning},
  author={Habibian, Soheil and Jonnavittula, Ananth and Losey, Dylan P},
  journal={ACM Transactions on Human-Robot Interaction},
  year={2022},
  publisher={ACM New York, NY}
}

@article{harmon1997reinforcement,
  title={Reinforcement Learning: A Tutorial.},
  author={Harmon, Mance E and Harmon, Stephanie S},
  year={1997},
  publisher={Wright Lab Wright-Patterson Afb Oh}
}

@inproceedings{dagger,
  title={Efficient reductions for imitation learning},
  author={Ross, St{\'e}phane and Bagnell, Drew},
  booktitle={Proceedings of the thirteenth international conference on artificial intelligence and statistics},
  pages={661--668},
  year={2010},
  organization={JMLR Workshop and Conference Proceedings}
}

@inproceedings{drex,
  title={Better-than-demonstrator imitation learning via automatically-ranked demonstrations},
  author={Brown, Daniel S and Goo, Wonjoon and Niekum, Scott},
  booktitle={Conference on robot learning},
  pages={330--359},
  year={2020},
  organization={PMLR}
}

@inproceedings{trex,
  title={Extrapolating beyond suboptimal demonstrations via inverse reinforcement learning from observations},
  author={Brown, Daniel S. and Goo, Wonjoon and Nagarajan, Prabhat and Niekum, Scott},
  booktitle={International conference on machine learning},
  pages={783--792},
  year={2019},
  organization={PMLR}
}

@inproceedings{yu2020meta,
  title={Meta-world: A benchmark and evaluation for multi-task and meta reinforcement learning},
  author={Yu, Tianhe and Quillen, Deirdre and He, Zhanpeng and Julian, Ryan and Hausman, Karol and Finn, Chelsea and Levine, Sergey},
  booktitle={Conference on robot learning},
  pages={1094--1100},
  year={2020},
  organization={PMLR}
}

@article{spencer2022expert,
  title={Expert intervention learning},
  author={Spencer, Jonathan and Choudhury, Sanjiban and Barnes, Matthew and Schmittle, Matthew and Chiang, Mung and Ramadge, Peter and Srinivasa, Sidd},
  journal={Autonomous Robots},
  volume={46},
  number={1},
  pages={99--113},
  year={2022},
  publisher={Springer}
}

@inproceedings{bajcsy2018learning,
  title={Learning from physical human corrections, one feature at a time},
  author={Bajcsy, Andrea and Losey, Dylan P and O'Malley, Marcia K and Dragan, Anca D},
  booktitle={Proceedings of the 2018 ACM/IEEE International Conference on Human-Robot Interaction},
  pages={141--149},
  year={2018}
}

@article{biyik2019batch,
 title={Batch Active Learning Using Determinantal Point Processes},
 author={Biyik, Erdem and Wang, Kenneth and Anari, Nima and Sadigh, Dorsa},
 journal={arXiv preprint arXiv:1906.07975},
 year={2019}
}

@inproceedings{biyik2020active,
 title={Active Preference-Based Gaussian Process Regression for Reward Learning},
 author={Biyik, Erdem and Huynh, Nicolas and Kochenderfer, Mykel J. and Sadigh, Dorsa},
 booktitle={Proceedings of Robotics: Science and Systems (RSS)},
 year={2020},
 month=jul
}

@inproceedings{biyik2018batch,
  title={Batch active preference-based learning of reward functions},
  author={Biyik, Erdem and Sadigh, Dorsa},
  booktitle={Conference on robot learning},
  pages={519--528},
  year={2018},
  organization={PMLR}
}

@article{biyik2021learning,
 title={Learning Reward Functions from Diverse Sources of Human Feedback: Optimally Integrating Demonstrations and Preferences},
 author={Biyik, Erdem and Losey, Dylan P. and Palan, Malayandi and Landolfi, Nicholas C. and Shevchuk, Gleb and Sadigh, Dorsa},
 journal={The International Journal of Robotics Research (IJRR)},
 year={2021}
}

@article{grondman2012survey,
  title={A survey of actor-critic reinforcement learning: Standard and natural policy gradients},
  author={Grondman, Ivo and Busoniu, Lucian and Lopes, Gabriel AD and Babuska, Robert},
  journal={IEEE Transactions on Systems, Man, and Cybernetics, Part C (Applications and Reviews)},
  volume={42},
  number={6},
  pages={1291--1307},
  year={2012},
  publisher={IEEE}
}

@article{konda1999actor,
  title={Actor-critic algorithms},
  author={Konda, Vijay and Tsitsiklis, John},
  journal={Advances in neural information processing systems},
  volume={12},
  year={1999}
}

@article{haarnoja2018soft,
  title={Soft actor-critic algorithms and applications},
  author={Haarnoja, Tuomas and Zhou, Aurick and Hartikainen, Kristian and Tucker, George and Ha, Sehoon and Tan, Jie and Kumar, Vikash and Zhu, Henry and Gupta, Abhishek and Abbeel, Pieter and others},
  journal={arXiv preprint arXiv:1812.05905},
  year={2018}
}

@inproceedings{sadigh2017active,
 author = {Sadigh, Dorsa and Dragan, Anca D. and Sastry, S. Shankar and Seshia, Sanjit A.},
 title = {Active Preference-Based Learning of Reward Functions},
 booktitle = {Proceedings of Robotics: Science and Systems (RSS)},
 year = {2017},
 month = jul
}

@inproceedings{wilde2020active,
  author={N. {Wilde} and D. {Kulić} and S. L. {Smith}},
  booktitle={2020 IEEE/RSJ International Conference on Intelligent Robots and Systems (IROS)}, 
  title={Active Preference Learning using Maximum Regret}, 
  year={2020},
  pages={10952-10959}
}

@article{hui2020babyai,
  title={BabyAI 1.1},
  author={Hui, David Yu-Tung and Chevalier-Boisvert, Maxime and Bahdanau, Dzmitry and Bengio, Yoshua},
  journal={arXiv preprint arXiv:2007.12770},
  year={2020}
}

@article{mnih2013playing,
  title={Playing atari with deep reinforcement learning},
  author={Mnih, Volodymyr and Kavukcuoglu, Koray and Silver, David and Graves, Alex and Antonoglou, Ioannis and Wierstra, Daan and Riedmiller, Martin},
  journal={arXiv preprint arXiv:1312.5602},
  year={2013}
}

@article{wirth2017survey,
  title={A survey of preference-based reinforcement learning methods},
  author={Wirth, Christian and Akrour, Riad and Neumann, Gerhard and F{\"u}rnkranz, Johannes and others},
  journal={Journal of Machine Learning Research},
  volume={18},
  number={136},
  pages={1--46},
  year={2017},
  publisher={Journal of Machine Learning Research/Massachusetts Institute of Technology~…}
}

@inproceedings{li2021roial,
 title = {ROIAL: Region of Interest Active Learning for Characterizing Exoskeleton Gait Preference Landscapes},
 author = {Li, Kejun and Tucker, Maegan and Biyik, Erdem and Novoseller, Ellen and Burdick, Joel W. and Sui, Yanan and Sadigh, Dorsa and Yue, Yisong and Ames, Aaron D.},
 booktitle={International Conference on Robotics and Automation (ICRA)},
 year = {2021},
 month = may
}

@inproceedings{chen2013pairwise,
  title={Pairwise ranking aggregation in a crowdsourced setting},
  author={Chen, Xi and Bennett, Paul N and Collins-Thompson, Kevyn and Horvitz, Eric},
  booktitle={Proceedings of the sixth ACM international conference on Web search and data mining},
  pages={193--202},
  year={2013}
}

@inproceedings{chen2017near,
  title={Near-optimal active learning of halfspaces via query synthesis in the noisy setting},
  author={Chen, Lin and Hassani, Hamed and Karbasi, Amin},
  booktitle={Proceedings of the AAAI Conference on Artificial Intelligence},
  volume={31},
  number={1},
  year={2017}
}

@inproceedings{wise2016fetch,
  title={Fetch and freight: Standard platforms for service robot applications},
  author={Wise, Melonee and Ferguson, Michael and King, Derek and Diehr, Eric and Dymesich, David},
  booktitle={Workshop on autonomous mobile service robots},
  year={2016}
}

@inproceedings{cakmak2011human,
  title={Human preferences for robot-human hand-over configurations},
  author={Cakmak, Maya and Srinivasa, Siddhartha S and Lee, Min Kyung and Forlizzi, Jodi and Kiesler, Sara},
  booktitle={2011 IEEE/RSJ International Conference on Intelligent Robots and Systems},
  pages={1986--1993},
  year={2011}
}

@article{hindsight,
  title={Shared autonomy via hindsight optimization},
  author={Javdani, Shervin and Srinivasa, Siddhartha S and Bagnell, J Andrew},
  journal={Robotics science and systems: online proceedings},
  volume={2015},
  year={2015},
  publisher={NIH Public Access}
}

@inproceedings{da2020uncertainty,
  title={Uncertainty-aware action advising for deep reinforcement learning agents},
  author={Da Silva, Felipe Leno and Hernandez-Leal, Pablo and Kartal, Bilal and Taylor, Matthew E},
  booktitle={Proceedings of the AAAI Conference on Artificial Intelligence},
  volume={34},
  number={04},
  pages={5792--5799},
  year={2020}
}

@inproceedings{biyik2019asking,
 title={Asking Easy Questions: A User-Friendly Approach to Active Reward Learning},
 author={Biyik, Erdem and Palan, Malayandi and Landolfi, Nicholas C. and Losey, Dylan P. and Sadigh, Dorsa},
 booktitle={Proceedings of the 3rd Conference on Robot Learning (CoRL)},
 year={2019}
}

@inproceedings{viappiani2010optimal,
  title={Optimal Bayesian Recommendation Sets and Myopically Optimal Choice Query Sets.},
  author={Viappiani, Paolo and Boutilier, Craig},
  booktitle={NeurIPS},
  pages={2352--2360},
  year={2010}
}

@inproceedings{myers2022learning,
  title={Learning multimodal rewards from rankings},
  author={Myers, Vivek and Biyik, Erdem and Anari, Nima and Sadigh, Dorsa},
  booktitle={Conference on Robot Learning},
  pages={342--352},
  year={2022},
  organization={PMLR}
}

@article{akgun2012keyframe,
  title={Keyframe-based learning from demonstration},
  author={Akgun, Baris and Cakmak, Maya and Jiang, Karl and Thomaz, Andrea L},
  journal={International Journal of Social Robotics},
  volume={4},
  number={4},
  pages={343--355},
  year={2012},
  publisher={Springer}
}

@misc{highway-env,
  author = {Leurent, Edouard},
  title = {An Environment for Autonomous Driving Decision-Making},
  year = {2018},
  publisher = {GitHub},
  journal = {GitHub repository},
  howpublished = {\url{https://github.com/eleurent/highway-env}},
}

@inproceedings{hejna2022fewshot,
 title={Few-Shot Preference Learning for Human-in-the-Loop RL},
 author={Hejna, III, Donald Joseph and Sadigh, Dorsa},
 booktitle={Proceedings of the 6th Conference on Robot Learning (CoRL)},
 year={2022}
}

@article{jeon2020reward,
  title={Reward-rational (implicit) choice: A unifying formalism for reward learning},
  author={Jeon, Hong Jun and Milli, Smitha and Dragan, Anca},
  journal={Advances in Neural Information Processing Systems},
  volume={33},
  pages={4415--4426},
  year={2020}
}

@article{stable-baselines3,
  author  = {Antonin Raffin and Ashley Hill and Adam Gleave and Anssi Kanervisto and Maximilian Ernestus and Noah Dormann},
  title   = {Stable-Baselines3: Reliable Reinforcement Learning Implementations},
  journal = {Journal of Machine Learning Research},
  year    = {2021},
  volume  = {22},
  number  = {268},
  pages   = {1-8}
}

@inproceedings{hoque2021thriftydagger,
  title={ThriftyDAgger: Budget-Aware Novelty and Risk Gating for Interactive Imitation Learning},
  author={Hoque, Ryan and Balakrishna, Ashwin and Novoseller, Ellen and Wilcox, Albert and Brown, Daniel S and Goldberg, Ken},
  booktitle={5th Annual Conference on Robot Learning},
  year={2021}
}

@inproceedings{kelly2019hg,
  title={Hg-dagger: Interactive imitation learning with human experts},
  author={Kelly, Michael and Sidrane, Chelsea and Driggs-Campbell, Katherine and Kochenderfer, Mykel J},
  booktitle={2019 International Conference on Robotics and Automation (ICRA)},
  pages={8077--8083},
  year={2019},
  organization={IEEE}
}

\newpage
\onecolumn
\appendix
\let\section\subsection
\let\subsection\subsubsection

In the appendix, we present details about our approach and experiments. See \url{http://tinyurl.com/online-active} for a website summarizing our results.

\section{Derivation}
\label{app:derivation}
\def\E{\mathop{\mathbb E}}

We adopt the setting in \Cref{sec:approach}. We define the set of past query responses to be $\D=\Bigl(\bigl(a^{(1)}_{i_1}, s\t 1, (a^{(1)}_{1}, a^{(1)}_{2})\bigr),\bigl(a^{(2)}_{i_2},s\t 2, (a^{(2)}_{1}, a^{(2)}_{2})\bigr),\ldots, \bigl(a^{(n)}_{i_n},s\t n, (a^{(n)}_{1}, a^{(n)}_{2})\bigr)\Bigr)$ for each $i_\bullet \in \left\{1,2\right\}$, and denote $\D_1=\D\cup\left\{(a_1, s, (a_1, a_2))\right\},$ $\D_2= \D\cup \left\{(a_2, s, (a_1, a_2))\right\}$ for fixed actions $(a_1, a_2)$ to potentially query in state $s$. The EVOI is defined as the expected change in the value function $V(s;\omega)=\max_a Q^*(s,a;\omega)$ of the current state after asking a query and updating our belief state about the policy $Q$ function.
Assuming optimality of our learned $\max_a Q^\pi(s,a;\omega)$, we obtain the following formula for the EVOI of the query $(a_1,a_2)$.

\begin{align*}
    \operatorname{EVOI}(a_1, a_2) 
    &= \E_{\text{query response}}\left[ \E_{\omega\mid \text{after query}} V(s;\omega)\right] -\E_{\omega\mid \text{before query}} V(s;\omega) 
    \\&=\biggl[\E_{\omega\mid\Omega,\D} H^{\pi^*}(a_1, a_2, s, \omega)\biggr] \E_{\omega\mid \Omega,\D_1} V(s;\omega) + \biggl[\E_{\omega\mid\Omega,\D} H^{\pi^*}(a_2, a_1, s, \omega)\biggr]\E_{\omega\mid \Omega,\D_2} V(s;\omega) - \E_{\omega\mid\Omega,\D}V(s. ; \omega)
     \\&=\E_{\omega\mid\Omega,\D}\Bigl[ H^\pi(a_1, a_2, s, \omega) \E_{\omega'\mid \Omega,\D_1} [V(s;\omega')]+ H^\pi(a_2, a_1, s, \omega)  \E_{\omega'\mid \Omega,\D_2} [V(s;\omega')]-V(s; \omega)\Bigr]
     \\&=\E_{\omega\mid\Omega,\D}\Bigl[ H^\pi(a_1, a_2, s, \omega) \E_{\omega'\mid \Omega,\D_1} \max_a Q^\pi(s,a;\omega') \\&\qquad\qquad
     + H^\pi(a_2, a_1, s, \omega)  \E_{\omega'\mid \Omega,\D_2} \max_a Q^\pi(s,a;\omega')-\max_a Q^\pi(s, a; \omega)\Bigr]
\end{align*}

\begin{revision}
\section{GridWorld Experiments}
\label{app:grid}

\begin{figure}[hbt]
\centering
  \includegraphics[width=0.9\linewidth]{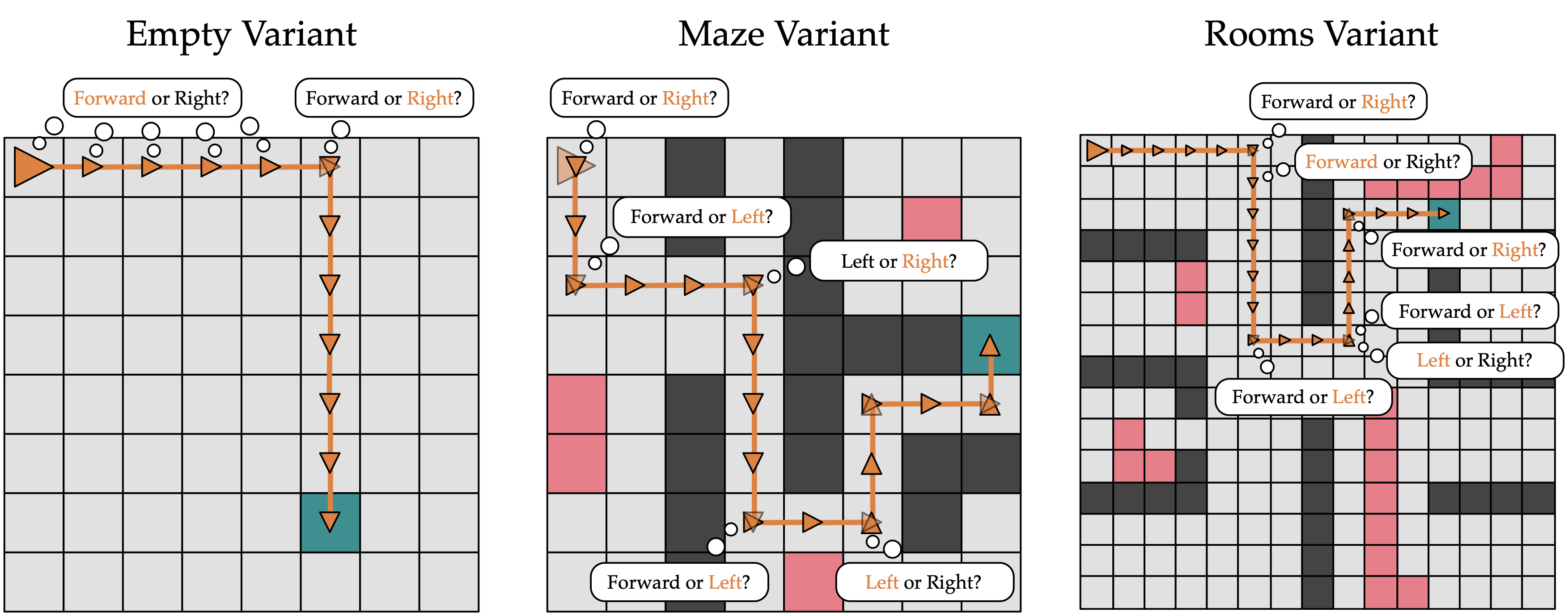}
  \caption{GridWorld variants studied. Agent start position and orientation are shown by the large orange arrow. Empty spaces are gray, walls are black, and lava is red. The goal location is randomized uniformly across empty spaces at the start of each episode. In the diagrams above, we show a trajectory of our approach on each environment for a randomly-selected goal location, as well as the queries made by our method (with expert responses highlighted in orange).}
  \label{fig:grids}
\end{figure}

\begin{figure}[htb]
    \centering
    \includegraphics[width=.55\linewidth]{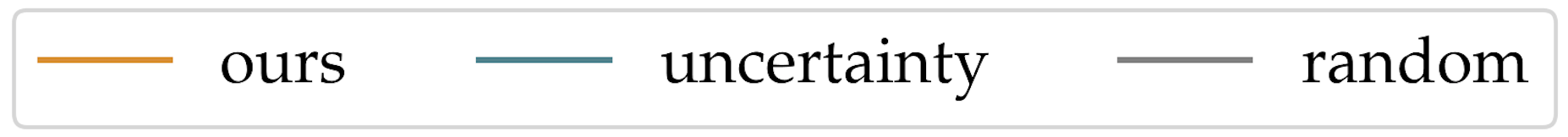}\par
    \includegraphics[width=.3333\linewidth]{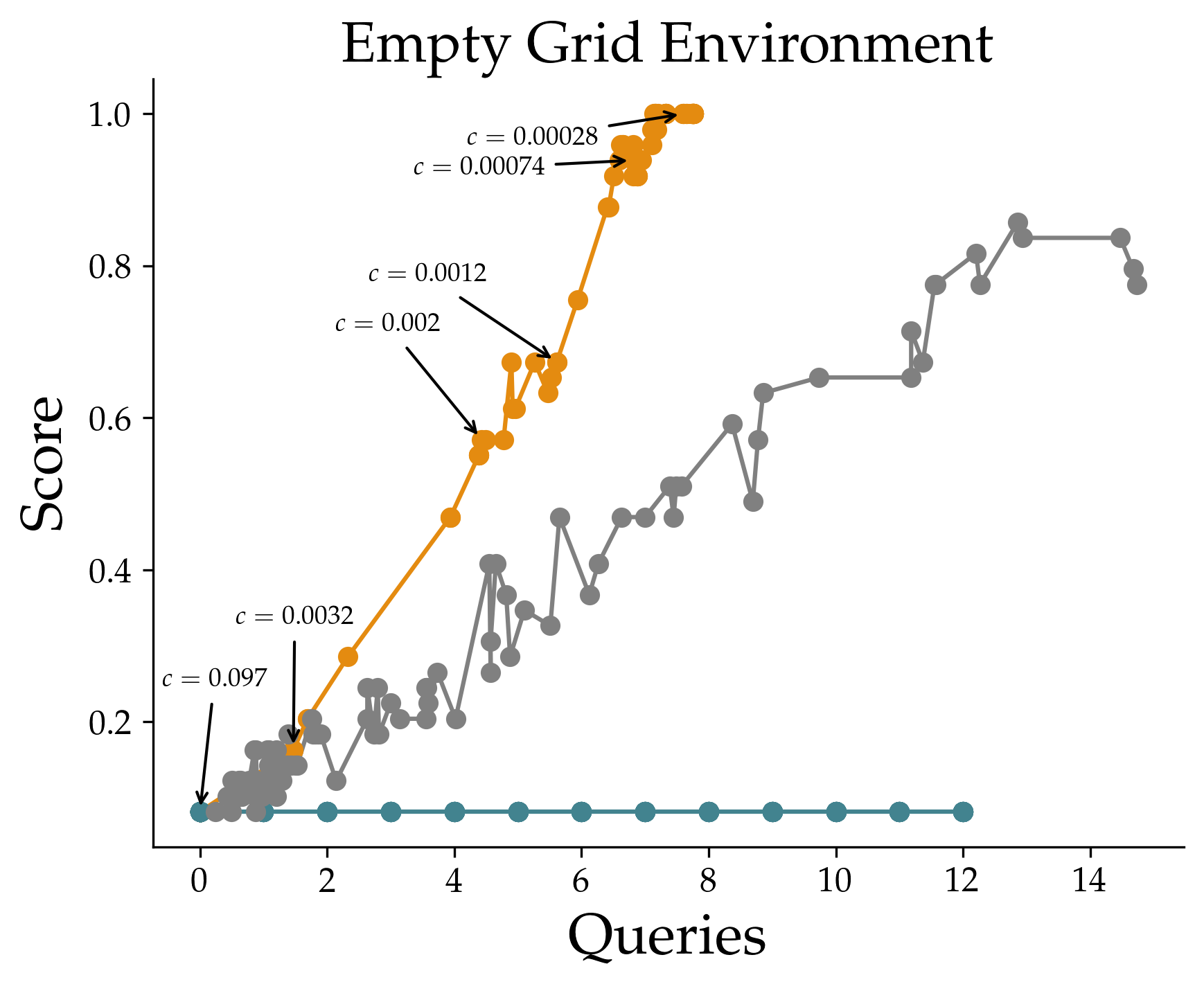}%
    \includegraphics[width=.3333\linewidth]{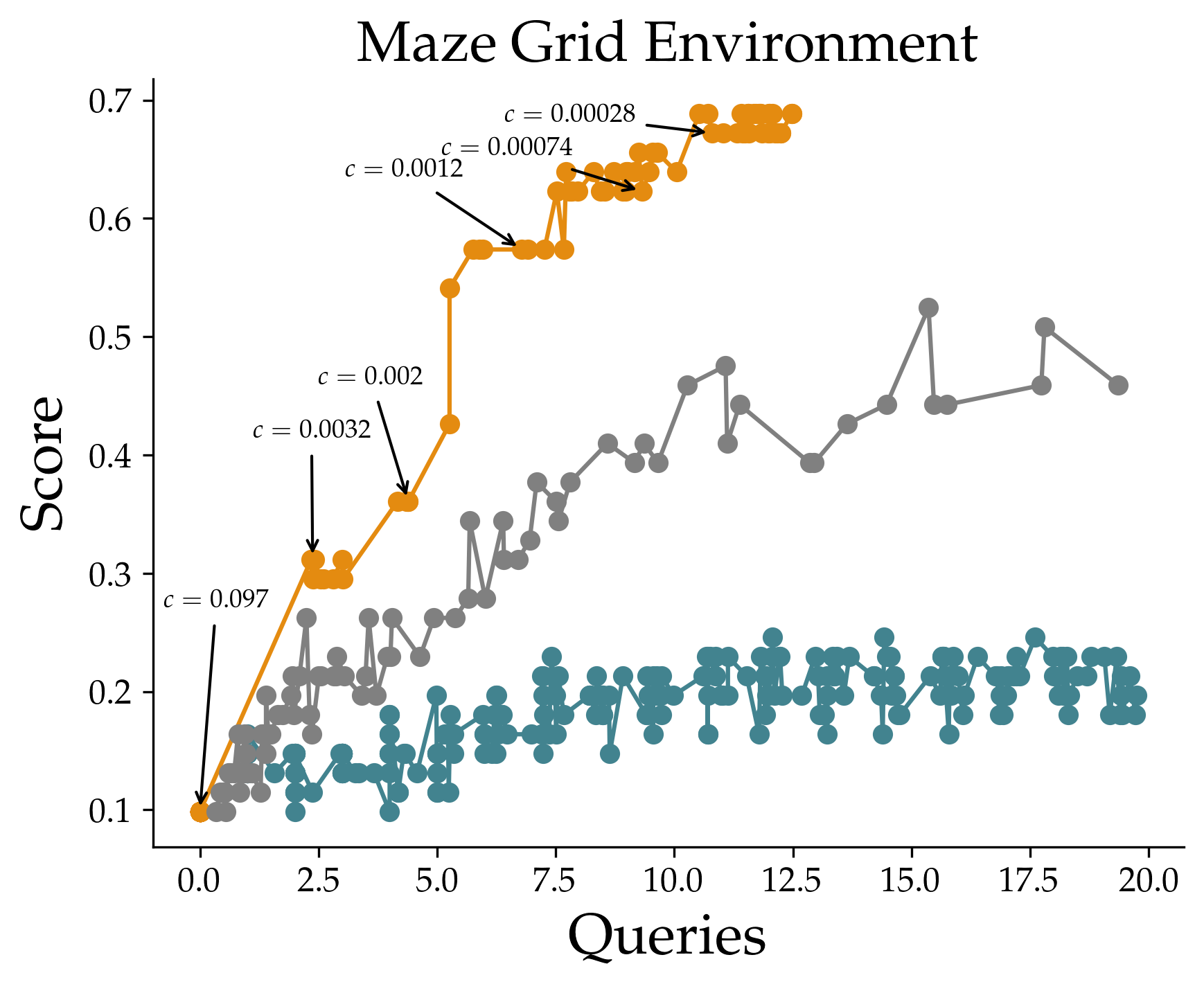}%
    \includegraphics[width=.3333\linewidth]{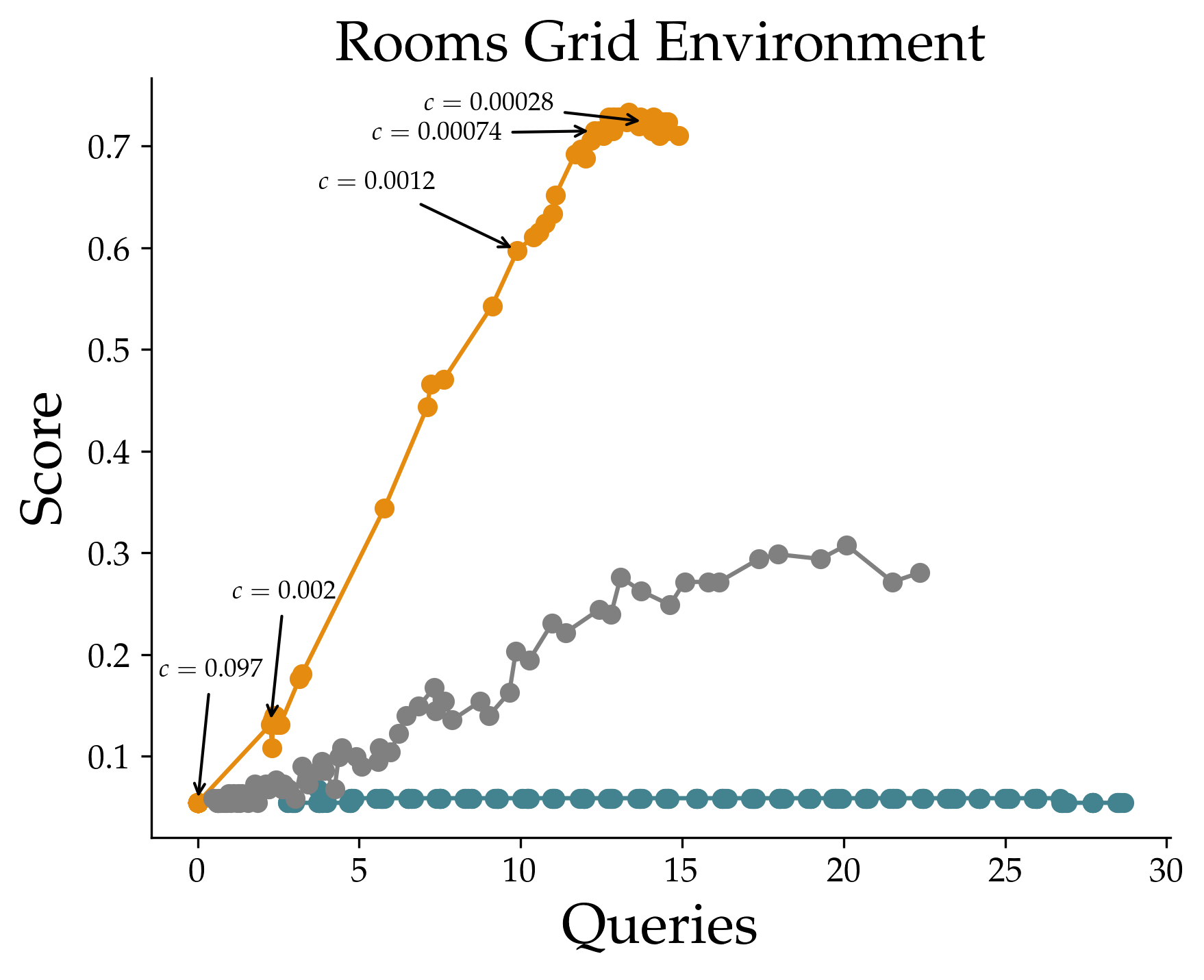}
    \caption{Pareto-frontiers of methods on the GridWorld environment variants. Each point in the plots represents the number of queries made and average score of running a method for a particular initialization of querying parameters.  By varying the querying thresholds for each approach, we see the trade offs between score and number of questions asked. For our approach, we additionally label the values of $c$ used to generate some points to compare robustness to different $c$ values across GridWorld settings.
    }
    \label{fig:gridresults}
\end{figure}

Our initial results test our algorithm in a GridWorld environment \cite{hui2020babyai}. This environment involves an agent navigating a grid to reach a goal destination. In this setting, our task distribution $\Omega$ is a uniform distribution over possible valid goal locations. We train a policy conditioned on goal locations $\omega\in\Omega$ to navigate these environments using a tabular model trained with value iteration \cite{harmon1997reinforcement}. We then compare our querying approach against the baseline approaches described above, using the learned policy conditioned on the true goal as our expert policy. We focus on the following three GridWorld environment variants, shown in \Cref{fig:grids}:
\textbf{Empty Variant:} The agent starts in the upper left corner of a 8x8 GridWorld and navigates to a random goal location within the grid. 
\textbf{Maze Variant:} The agent starts in the upper left corner of a 8x8 GridWorld and navigates to a random goal location within the grid, avoiding a maze of walls and lava. 
\textbf{Rooms Variant:} The agent starts in the upper left corner of a 15x15 GridWorld environment, partitioned into multiple room sections with walls and lava, and navigates to a random grid location.

We evaluate our approach against baselines in these environments by comparing the Pareto frontiers of their average performance in terms of score achieved and average number of queries made per trajectory. We vary the expected number of queries made by our approach and baselines by varying the querying parameters of each method ($c$ for our approach). By plotting the Pareto frontiers of our approach against baselines, we can compare for a fixed number of queries made by an approach, the expected score achieved in an environment. These Pareto frontiers are presented in \Cref{fig:gridresults}. 

Our approach generally outperforms both baseline approaches, achieving higher score (y-axis value) for given average numbers of questions asked (x-axis value). Interestingly, the uncertainty baseline performs significantly worse than the random baseline in this setting. This performance is likely due to to the fact that in a GridWorld trajectory, even though there is high uncertainty over the goal location, all policies are able to get roughly similar $Q$ values since the environment is fully solvable, resulting in queries that may be strongly disconnected from imp ortant decision points. Since the uncertainty querying is deterministic, unlike random, it can then get in states where it needs to ask a question but never will, resulting in very poor performance. 

We additionally label some of the points in the Pareto frontier for our approach with the $c$ values used to generate them. By comparing the location of the points with the same value of $c$ across GridWorld settings we can see the value of $c$ transfers across similar settings. For instance, in all three settings, once $c$ drops below $\approx\!0.0012$ the agent starts to exceed a score of $0.5$ and use $5$-$10$ queries. Thus, similar $c$ values tend to result in similar querying behavior and performance across different GridWorld settings.

\subsection{Details}
\end{revision}

The action space in the GridWorld environment contains three actions: $A = \{\text{turn left}, \text{turn right}, \text{move forward}\}$. The agent cannot move through wall spaces, and the episode immediately ends upon moving into a lava square. To compute parteo-frontiers, we vary the querying parameters for our three approaches as follows:
\begin{description}
\item[Ours:] parameter varies from $10^{-4}$ to $10^{-1}$ with a step size of $\log(1.05)$ in log space.
\item[Uncertainty:] parameter varies from $10^{-4}$ to $10^{1}$ with a step size of $\log(1.05)$ in log space.
\item[Random:] parameter varies from $0.05$ to $0.5$ with a step size of $\log(1.05)$ in log space.
\end{description}

Across experiments, we use a human response precision parameter of $\beta=10$. We used value iteration to directly solve for optimal policies across tasks, using $\gamma=0.99$ and a fixed horizon of 50 steps. These and all experiments in this paper were run using \texttt{Intel(R) Xeon(R) Silver 4114 CPU @ 2.20GHz} processors.

\section{Driving Details}
\label{app:driving}

We use a human response precision parameter of $\beta=10$ across all experiments. For the the driving environment experiments in \Cref{sec:drivexp}, we use $c=0.05$ for our approach, a querying probability of $0.2$ for the random baseline, and an uncertainty threshold of $46$ for the uncertainty baseline. 
These hyperparameters were tuned so that the approaches ask similar numbers of questions to better facilitate comparisons of our approach against baselines. 

Our DQNs for use by agents and expert policies were trained across a random distribution of tasks, with 100 tasks with trained DQN policies used for controlling agents and the remaining 487 tasks and trained DQN policies used as evaluation tasks and human expert simulators respectively. We trained our DQNs for 60,000 steps with a batch size of 32, a replay buffer of size 15,000, an Adam learning rate of $5\cdot 10^{-4}$, $\gamma=0.8$, and a two-hidden-layer, 256 unit MLP architecture.

\subsection{Pareto Frontiers}
\label{app:Pareto-driving}

We additionally extend the driving experiments to construct full Pareto frontiers modelling the full relationship between the number of questions asked and the score achieved by each method. As in \Cref{app:grid}, we vary the querying parameter of each approach to generate points on this frontier. The resulting frontier is presented in \Cref{fig:pareto-combined}. We see our approach outperforms baselines in terms of score across different numbers of queries made.

We varied the querying parameters of the approaches over the following values:
\begin{description}
\item[Ours:] 0.002, 0.01, 0.05, 0.25, and 1.25.
\item[Uncertainty:] 30.67, 46, 69, 103.5, and 155.25.
\item[Random:] 0.025, 0.05, 0.1, 0.2, and 0.4.
\end{description}

\section{Block Pushing Details}
\label{app:fetch}

We use a human response precision parameter of $\beta=10$, a threshold of $c=0.03$ for our approach, a querying probability of $0.25$ for the random baseline, and an uncertainty threshold of $0.01$ for the uncertainty baseline across the Fetch block pushing experiments.

We trained expert policies using soft actor-critic for the block pushing task. We trained these policies in a simulated Fetch block pushing environment \cite{plappert2018multi}. To apply our approaches to this setting, we used the generalization to continuous action spaces described in \Cref{sec:continuous}.

We learned a general policy parameterized by the block's current position as well as the goal position using soft actor-critic with a 2-hidden-layer 256 unit MLP architecture. These policies required the current location of the block to be in the environments observation space, so we used the simulator to keep track to model the block's location when deploying our approaches.

Using waypoints generated by the simulator, we transferred trajectories from simulation to the real Fetch robot.

\subsection{Pareto Frontiers}
\label{app:Pareto-pushing}

We allowed querying parameters to vary over the following values for each approach to generate this plot:

\begin{description}
\item[Ours:] 0.0075, 0.015, 0.03, 0.06, and 0.12.
\item[Uncertainty:] 0.0025, 0.005, 0.01, 0.02, and 0.04.
\item[Random:] 0.111, 0.167, 0.25, 0.375, and 0.563.
\end{description}

\end{document}